\RequirePackage{fix-cm}
\documentclass[twocolumn]{svjour3}          
\smartqed  
\usepackage{graphicx}

\usepackage{amssymb}
\usepackage{comment}
\usepackage[multiple]{footmisc}
\usepackage{amsmath}
\usepackage{kbordermatrix}
\usepackage{lipsum}
\usepackage{pfnote}
\usepackage{multirow}
\usepackage{float}
\usepackage{pgfplots}
\usepackage{tikz}
\usepackage{xcolor}
\usetikzlibrary{patterns}
\usepackage{amstext}
\usepackage{algorithm} 
\usepackage{algpseudocode} 
\usetikzlibrary{arrows,decorations.pathmorphing,fit,positioning}

\usepackage[hyphenbreaks]{breakurl}
\usepackage[hyphens]{url}

\usepackage{mathtools}

\usetikzlibrary{positioning,calc}

\def\addlegendimage{\csname pgfplots@addlegendimage\endcsname}

\usepackage[square,numbers]{natbib}

\usetikzlibrary{positioning,calc}

\newenvironment{customlegend}[1][]{%
	\begingroup
	\csname pgfplots@init@cleared@structures\endcsname
	\pgfplotsset{#1}%
}{%
\csname pgfplots@createlegend\endcsname
\endgroup
}%

\def\addlegendimage{\csname pgfplots@addlegendimage\endcsname}

\begin{document}
	
	\title{Fuzzy Approach Topic Discovery in Health and Medical  Corpora
	}
	
	
	\author{Amir Karami         \and
		Aryya Gangopadhyay \and
		Bin Zhou \and
		Hadi Kharrazi
	}
	
	
	\institute{Amir Karami \at
		School of Library and Information Science, University of South Carolina \\
		Tel.: +1-803-777-0197\\
		Fax: +1-803-777-7938\\
		\email{karami@sc.edu}           
		\and
		Aryya Gangopadhyay and Bin Zhou \at
		Information Systems Department, University of Maryland Baltimore County 
		\and
		Hadi Kharrazi\at
		Bloomberge School of Public Health, Johns Hopkins University 
	}
	
	\date{Received: date / Accepted: date}

	\maketitle
	
	\begin{abstract}
			The majority of medical documents and electronic health records (EHRs) are in text format that poses a challenge for data processing and finding relevant documents. Looking for ways to automatically retrieve the enormous amount of health and medical knowledge has always been an intriguing topic. Powerful methods have been developed in recent years to make the text processing automatic. One of the popular approaches to retrieve information based on discovering the themes in health \& medical corpora is topic modeling; however, this approach still needs new perspectives. In this research we describe \textit{fuzzy latent semantic analysis} (FLSA), a novel approach in topic modeling using fuzzy perspective. FLSA can handle health \& medical corpora redundancy issue and provides a new method to estimate the number of topics. The quantitative evaluations show that FLSA produces superior performance and features to \textit{latent Dirichlet allocation} (LDA), the most popular topic model.
	\end{abstract}
	
\section{Introduction}
\label{Int}
There is a growing need to analyze large collections of electronic documents. Moreover, very large-scale scientific data management and analysis is one of the data-intensive challenges identified by National Science Foundation (NSF) as an area for future study~\cite{council2016future}. Large collections of electronic documents abound as our collective knowledge continues to be digitized and stored, requiring new tools for organization, search, indexing, and browsing. As a consequence, finding relevant documents has become more difficult for experts. In particular, large scale health and medical text data historically has been generated and stored. For example, the total number of papers published on PubMed website is more than 6 million papers in 2015\footnote{\url{http://www.ncbi.nlm.nih.gov/pubmed}} and the annual average number of US hospital discharges is more than 30 million records \cite{karami2015flatm,karami2015fuzzyiconf}. This huge amount of text data and EHRs is a great motivation for companies to save \$450 billion a year using advanced data analytical approaches\footnote{\url{http://www.mckinsey.com/industries/healthcare-systems-and-services/our-insights/the-big-data-revolution-in-us-health-care}}. 

Developing efficient techniques for discovering the hidden structure in large complicated health and medical data sets, and using that structure to answer questions about those data, is at the core of big health and medical data science research. Substantial resources were allocated in developing new data analytic methods and tools. However, retrieving big health and medical text data is a major current challenge. 

One of the popular methods in medical text data representation is bag-of-words (BOW). This technique represents documents based on the frequency of words with a matrix like $A$. 
\[
A= \kbordermatrix{
	& Word_1 & Word_2 & Word_3 & Word_4 \\
	Document_1 & 3 & 1 & 4 & 0  \\
	Document_2 & 0 & 1 & 0 & 0  \\
	Document_3 & 0 & 0 & 3 & 0  \\ 	
}
\]

For example, matrix $A$ shows that word 3 appeared 3 times in document 3. However, this matrix is a sparse matrix for large number of documents \cite{karami2014fftm}. Sparsity means that there are a lot of words in a corpus; however, one document covers a small percentage of all words. Therefore, most elements are zero in BOW matrix \cite{aggarwal2012introduction}. 

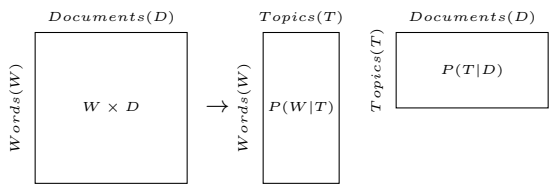
\begin{figure}[H]
	\centering
\begin{tikzpicture}
\tiny
\draw (-0.25,0) node [rotate=90] {$Words (W)$};
\draw (1,1.2) node {$Documents (D)$};
\draw (3.5,1.2) node {$Topics (T)$};
\draw (5.75,1.2) node {$Documents (D)$};
\draw (2.75,0) node [rotate=90] {$Words (W)$};
\draw (4.5,0.5) node [rotate=90] {$Topics (T)$};

\draw (0,-1) rectangle (2,1) node[pos=.5] {$W \times D$};
\small \draw (2.4,0) node {$\rightarrow$};
\tiny
\draw (3,-1) rectangle (4,1) node[pos=.5] {$P(W|T)$};
\draw (4.75,0) rectangle (6.75,1) node[pos=.5] {$P(T|D)$};


\end{tikzpicture} 
\caption{Matrix Interpretation of Topic Modeling}
\label{tab: drtm}
\end{figure}

Topic modeling is a popular method to address sparsity and high dimensionality issues. This method was originally introduced as a text analysis technique that the objects are documents and the features are the frequency of terms. The output of topic modeling is two matrices. The first one is the probability of words for each topic or $P(W|T)$ and the second one is the probability of topics for each document or $P(T|D)$ (Figure~\ref{tab: drtm}). 

The words with higher probability in $Words \times Topic$ matrix discloses semantic structure. $Topics \times Documents$ matrix reduces the number of dimension from the number of words in BOW approach to the number of topics. For example, suppose that there are 100 topics in a corpus with 5000 documents and 10,000 words. Topic modeling converts $W \times D$ matrix to two matrices: Words $\times$ Topics matrix with 10,000 rows and 100 columns, and Topics $\times$ Documents matrix with 100 rows and 5000 columns. It is worth mentioning that the second non-sparse matrix is used for document classification and clustering with 100 topics as the number of features instead of 10,000 words as the number of features in the BOW approach. Document classification and clustering problems categorized labeled and unlabeled documents based on extracted features (topics) from documents using topic modeling.

Topic modeling is an effective method for health and medical text mining; however, due to the intensive amount of available data, there is still the need to improve the performance of this approach. In addition, copy and paste~(redundancy) has a negative impact on topic modeling \cite{cohen2013redundancy} and previous work has shown that most of medical notes are redundant \cite{wrenn2010quantifying}.

In this research, we propose  \textit{\underline{F}uzzy \underline{L}atent \underline{S}emantic \underline{A}nalysis} (FLSA) model for health and medical text mining. This model shows better performance in both redundant and non-redundant document and can help topic models estimating number of topics in corpus. The remainder of this paper is organized as follows. In the related work section, we review the related research. In the methodology section, we provide more details for FLSA. An empirical study was conducted to verify the effectiveness of FLSA. Finally, we provide an illustrative example for FLSA, and present a summary and future directions in the last two sections.

\section{Related Work}
\label{RW}

Text mining can be defined as the methods of machine learning and statistics with the goal of recognizing patterns and disclosing the hidden information in text data \cite{hotho2005brief}. In this section, we review key concepts, and health and medical applications of topic modeling and fuzzy clustering (FC).

		\vspace{-5mm}
\subsection{Topic Modeling in Health and Medical}

There are two main approaches in text mining: supervised and unsupervised. The goal of supervised approach is to disclose hidden structure in labeled datasets and the goal of unsupervised approach is to discover patterns in unlabeled datasets. The most popular techniques in supervised and unsupervised approaches are classification and clustering, correspondingly. The purpose of classification is to train a corpus with predefined labels and assign a label to a new document \cite{mitchell1997machine}. Clustering assigns a cluster to each document in a corpus based on similarity in a cluster and dissimilarity between clusters. Among text mining techniques, topic modeling is one of popular unsupervised methods with a wide range of applications from SMS spam detection \cite{karami2014exploiting} to image tagging \cite{xu2009tag}.

Topic modeling defines each topic as probability distribution over words and a document as probability distribution over topics. In health and medical text mining, latent Dirichlet allocation (LDA) shows better performance than other topic models \cite{sarioglu2012clinical}. 

		\vspace{-5mm}
\subsection{Health and Medical Applications of LDA}

LDA is a generative probabilistic model based on a three-level hierarchical Bayesian model. LDA assumes that documents contain latent topics and each topic can be represented by a distribution across words \cite{blei2003latent}.

LDA has a wide range of health and medical applications such as predicting protein-protein relationships based on the literature knowledge \cite{asou2008predicting}, discovering relevant clinical concepts and structures in patients' health records \cite{arnold2010clinical}, identifying patterns of clinical events in a cohort of brain cancer patients \cite{arnold2012topic}, and analyzing time-to-event outcomes \cite{dawson2012survival}. The discovery of clinical pathway (CP) patterns is a method for revealing the structure, semantics, and dynamics of CPs to provide clinicians with explicit knowledge used to guide treatment activities of individual patients. LDA has used for CPs to find treatment behaviors of patients \cite{huang2014similarity}, to predict clinical order patterns, and to model various treatment activities \cite{chen2016predicting} and their occurring time stamps in CPs \cite{defossez2014temporal}. LDA has also customized to determine patient mortality \cite{ghassemi2014unfolding}, and to discover knowledge from modeling disease and patient characteristics \cite{pivovarov2015learning}. Redundancy-aware LDA (Red-LDA) is one of the versions of LDA for handling redundancy issue in medical documents and has shown better performance than LDA \cite{cohen2014redundancy}.

\subsection{Health and Medical Applications of FC}

There are two major clustering approaches: hard and fuzzy~(soft). In hard clustering, every object may belong to exactly one cluster but, in fuzzy clustering (FC), the membership is fuzzy and objects may belong to several clusters \cite{karami2015fuzzy}. Among fuzzy clustering techniques, fuzzy C-means (FCM) is the most popular model \cite{bezdek1981pattern}. FCM is based on minimizing the overall distance from a cluster prototype to each datum.

Fuzzy clustering has used in predicting the response to treatment with citalopram in alcohol dependence \cite{naranjo1997using}, analyzing diabetic neuropathy \cite{di2002fuzzy}, detecting early diabetic retinopathy \cite{zahlmann2000hybrid}, characterizing stroke subtypes and coexisting causes of ischemic stroke \cite{helgason1998fuzzy,helgason2001statistical,helgason1999causal}, improving decision-making in radiation therapy \cite{papageorgiou2003integrated}, and detecting cancer such as breast cancer \cite{hassanien2003intelligent}. In addition, fuzzy clustering was used to improve ultrasound imaging technique \cite{moon2011breast} and analyze microarray data \cite{gasch2002exploring}. 

Although there are a lot of fuzzy clustering applications in health and medical domains especially in image processing, this approach has not been considered for topic modeling yet. This paper proposes a new approach to provide a bridge between fuzzy clustering and topic modeling to analyze big health and medical corpora.

\section{Methodology}
\label{Me}

In this part, we describe our method, \textit{fuzzy latent semantic analysis} (FLSA), for uncovering latent semantic features from text documents. FLSA treats fuzzy view as a new approach in topic modeling and  will be validated through a series of experiments, conducted on health and medical text data. 

Although LDA has shown a better performance than other topic models, redundancy has negative effect on LDA performance \cite{cohen2013redundancy,downey2006probabilistic}. The reason is that, for example, words like $w_1w_2w_3$ in a document such as $d_k=\{w_1w_2w_3w_6w_9\}$ are copied to a document like $d_p=\{w_7w_8\}$ to be $d_p=\{w_1w_2w_3w_7w_8\}$. $w_1w_2w_3$ should be assigned to the same topic by LDA but it is possible to be assigned by LDA to different topics. FLSA has a potential to handle the redundancy issue, estimate the optimum number of topics, and provide better performance than its competitors.  

		\vspace{-5mm}
\subsection{Fuzzy Logic and Fuzzy Clustering}
\begin{figure}[H]

	\centering
	\scalebox{0.5}{

		\begin{tikzpicture}[
		observed-1/.style={minimum size=50pt,circle,draw=black!10,fill=black!10},
		observed-2/.style={minimum size=50pt,circle,draw=black!15,fill=black!15},
		observed-3/.style={minimum size=50pt,circle,draw=black!20,fill=black!20},
		observed-4/.style={minimum size=50pt,circle,draw=black!25,fill=black!25},
		observed-5/.style={minimum size=50pt,circle,draw=black!30,fill=black!30},
		observed-6/.style={minimum size=50pt,circle,draw=black!35,fill=black!35},
		observed-7/.style={minimum size=50pt,circle,draw=black!40,fill=black!40},
		observed-8/.style={minimum size=50pt,circle,draw=black!45,fill=black!45},
		observed-9/.style={minimum size=50pt,circle,draw=black!50,fill=black!50},
		observed-10/.style={minimum size=50pt,circle,draw=black!0,fill=black!0},
		unobserved/.style={minimum size=90pt,circle,draw},
		hyper/.style={minimum size=1pt,circle,fill=black},
		post/.style={->,>=stealth',semithick},
		]
		\node(A) {
			\begin{tikzpicture}
			\Large
			\node (w-j-9) [observed-9] at (10.8,0) {\textbf{0.9}};
			\node (w-j-8) [observed-8] at (9,0) {\textbf{0.8}};
			\node (w-j-7) [observed-7] at (7.2,0) {\textbf{0.7}};
			\node (w-j-6) [observed-6] at (5.4,0) {\textbf{0.6}};
			\node (w-j-5) [observed-5] at (3.6,0) {\textbf{0.5}};
			\node (w-j-4) [observed-4] at (1.8,0) {\textbf{0.4}};
			\node (w-j-3) [observed-3] at (0,0) {\textbf{0.3}};
			\node (w-j-2) [observed-2] at (-1.8,0) {\textbf{0.2}};
			\node (w-j-1) [observed-1] at (-3.6,0) {\textbf{0.1}};
			
			\end{tikzpicture}
		};
		
	\end{tikzpicture}}

	\caption{Fuzzy Logic Spectrum}
	\label{fig:fcon}
\end{figure}
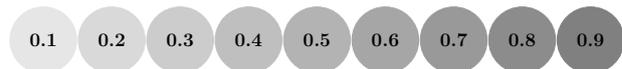

The traditional reasoning has precise character that is yes-or-no~(true-or-false) rather than more-or-less \cite{zimmermann2010fuzzy}. Fuzzy logic added a new extension to move from the classical logic, 0 or 1, to the truth values between zero and one, [0,1]) \cite{zadeh1973outline}~(Figure~\ref{fig:fcon}).

Fuzzy logic assumes that if $X$ is a collection of data points represented by $x$, then a fuzzy set $A$ in $X$ is a set of order pairs, $ A=\{(x,\mu_A (x)|x \in X)\}$.

$\mu_A(x)$ is the membership function which maps $X$ to the membership space $M$ which is between 0 and 1 \cite{karami2012fuzzy}. The goal of most clustering algorithms is to minimize the objective function $J$ that measures the quality of clusters to find the optimum $J$ which is the sum of the squared distances between each cluster center and each data point. 

The main goal of fuzzy models is to formulate uncertainty for applications such as decision-making \cite{karami2010risk,karami2012fuzzy}. For example, a voter decides to select some candidates among a set of candidates in an election. The voter has different preferences in terms of economic, foreign policy, health, etc. Based on the preferences, the distance between each candidate's plans and the voter's preferences can be changed. These preferences can be formulated and measured in fuzzy clustering with $\mu$, degree of membership.

\begin{figure*}[ht]
\Large
	\centering
	\scalebox{0.55}{

		\begin{tikzpicture}[
		observed-1/.style={minimum size=15pt,circle,draw=white,fill=white},
		observed-2/.style={minimum size=15pt,circle,draw=white,fill=white},
		observed-3/.style={minimum size=15pt,circle,draw=white,fill=white},
		unobserved/.style={minimum size=15pt,circle,draw},
		hyper/.style={minimum size=1pt,circle,fill=black},
		post/.style={->,>=stealth',semithick},
		]
		\node(A) {
			\begin{tikzpicture}
			\Large
			\node (w-j-1) [observed-1] at (0,0) {\textbf{human}};
			\node (w-j-2) [observed-2] at (-3,0) {\textbf{genome}};
			\node (w-j-3) [observed-3] at (-6,0) {\textbf{dna}};
			
			\node (w-j-4) [observed-2] at (0,-4) {\textbf{disease}};
			\node (w-j-5) [observed-3] at (-3,-4) {\textbf{bacteria}};
			\node (w-j-6) [observed-1] at (-6,-4) {\textbf{infectious}};

			\node (w-j-7) [observed-3] at (0,-8) {\textbf{evolution}};
			\node (w-j-8) [observed-1] at (-3,-8) {\textbf{species}};
			\node (w-j-9) [observed-2] at (-6,-8) {\textbf{organisms}};

			\node [draw,fit=(w-j-1) (w-j-9), inner sep=40pt] (plate-context) {};
			\node [above right] at (plate-context.south west) {$D$};
			\node [draw,fit=(w-j-1) (w-j-9), inner sep=15pt] (plate-token) {};
			\node [above right] at (plate-token.south west) {$N$};
			\end{tikzpicture}
		};
		\node[right=of A] (B) {
			\begin{tikzpicture}[
			observed-1/.style={minimum size=60pt,circle,draw=black!10,fill=black!10},
			observed-2/.style={minimum size=60pt,circle,draw=black!45,fill=black!20},
			observed-3/.style={minimum size=60pt,circle,draw=black!65,fill=black!40},
			unobserved/.style={minimum size=60pt,circle,draw},
			hyper/.style={minimum size=35pt,circle,fill=black},
			post/.style={->,>=stealth',semithick},
			]
			
			\node (w-j-1) [observed-1] at (0,0) {\textbf{human}};
			\node (w-j-2) [observed-2] at (-3,0) {\textbf{genome}};
			\node (w-j-3) [observed-3] at (-6,0) {\textbf{dna}};
			
			\node (w-j-4) [observed-2] at (0,-4) {\textbf{disease}};
			\node (w-j-5) [observed-3] at (-3,-4) {\textbf{bacteria}};
			\node (w-j-6) [observed-1] at (-6,-4) {\textbf{infectious}};

			\node (w-j-7) [observed-3] at (0,-8) {\textbf{evolution}};
			\node (w-j-8) [observed-1] at (-3,-8) {\textbf{species}};
			\node (w-j-9) [observed-2] at (-6,-8) {\textbf{organisms}};

			\node [draw,fit=(w-j-1) (w-j-9), inner sep=40pt] (plate-context) {};
			\node [above right] at (plate-context.south west) {$Topics$};
			\node [draw,fit=(w-j-1) (w-j-3), inner sep=5pt] (plate-token) {};
			\node [above right] at (plate-token.south west) {$T_1$};
			\node [draw,fit=(w-j-4) (w-j-6), inner sep=5pt] (plate-token) {};
			\node [above right] at (plate-token.south west) {$T_2$};
			\node [draw,fit=(w-j-7) (w-j-9), inner sep=5pt] (plate-token) {};
			\node [above right] at (plate-token.south west) {$T_3$};

			\end{tikzpicture}
		};

		\draw[line width=6pt,black,->] (A) -- (B);

	\end{tikzpicture}}

	\caption{This topic model example correspends to the fuzzy process.}
	\label{fig:ldaex}
\end{figure*}
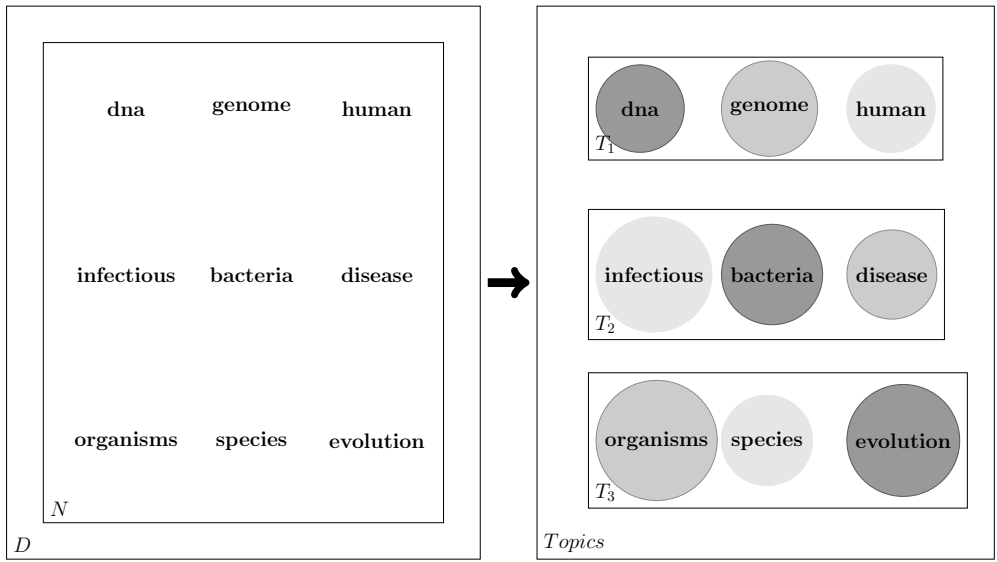

		\vspace{-5mm}

\subsection{FLSA}
FLSA assumes that documents and words can be fuzzy clustered and each cluster is a topic. For example, given a corpus FLSA discovers topic 1 including \textit{dna, genome, and human} words with ``Genetics" theme, topic 2 including \textit{infectious, bacteria, and disease} words with ``Disease" theme, and topic 3 including \textit{organisms, species, and evolution} words with ``Evolution" theme (Figure~\ref{fig:ldaex}). In this process, words are assigned a fuzzy degree of membership with respect to each cluster~(topic). The color of circles shows the magnitude of membership from low (light Grey) to high (dim Grey). 

The main goal of FLSA is to find two matrices: $P(T|D)$ and $P(W|T)$ mentioned in Figure \ref{tab: drtm}. FLSA has seven steps using \textit{Local Term Weighting (LTW), Global Term Weighting (GTM), and Fuzzy Clustering (FC)}:

\textbf{Step 1.} The first step is to calculate LTW. Among different LTW methods, we use term frequency (TF) as it is the most popular method.

\textbf{Step 2.} The next step is to calculate GTW. We explore four GTW methods in this research including \textit{Entropy, Inverse Document Frequency (IDF), Probabilistic Inverse Document Frequency (ProbIDF),} and \textit{Normal} (Table~\ref{tab:gtw}):

\begin{itemize}
	\itemsep0em 
	
	\item Entropy gives higher weight to the terms with less frequency in few documents \cite{dumais1992enhancing}.
	\item IDF assigns higher weights to rare terms and lower weights to common terms \cite{papineni2001inverse}.
	\item Normal is used to correct discrepancies in document lengths and also normalize the document vectors \cite{kolda1998limited}.
	\item ProbIDF is similar to IDF and assigns very low negative weight for the terms occurring in every document \cite{kolda1998limited}
	
\end{itemize}

Symbol $tf_{ij}$ defines the number of times word $i$ occurs in document $j$. With $m$ words and $n$ documents, we need to find $b(tf_{ij})$ and $p_{ij}$ for calculating the four mentioned GTW methods:

\begin{equation}
b(tf_{ij})=\left\{
\begin{array}{c l}     
1 & tf_{ij}>0\\
0 & tf_{ij}=0
\end{array}\right.
\end{equation}

\begin{equation}
p_{ij}=\frac{tf_{ij}}{\sum_j tf_{ij}}
\end{equation}

\begin{table}[H]
	\centering
	
	\scalebox{1}{
		\begin{tabular}{|p{1.4cm}|p{3.5cm}|} \hline
			\textbf{Name} & \textbf{Formula} \\ \hline
			Entropy & $1+ \frac{\sum_j p_{ij}\log_2 (p_{ij})}{\log_2 n}$ \\ \hline
			IDF & $\log_2 \frac{n}{\sum_j tf_{ij}}$   \\ \hline
			Normal & $\frac{1}{\sqrt{\sum_j tf_{ij}^2}}$ \\ \hline
			ProbIDF  & $\log_2 \frac{n-\sum_j b(tf_{ij})}{\sum_j b(tf_{ij})}$   \\ \hline

		\end{tabular}}
		\caption{GTW Methods}
		\label{tab:gtw}
	\end{table}

	The outputs of this step are the document term matrices with applied TF-Entropy, TF-IDF, TF-Normal, and TF-ProbIDF methods.\\
	
	\textbf{Step 3.}
We use Fuzzy C-means (FCM) in this research to fuzzy cluster the documented represented by the four mentioned GTW methods. FCM minimizes an objective function by considering constraints: 
	
	\begin{equation}
	Min \: \: J_q (\mu,V,X)=\sum_{k=1}^{c} \sum_{j=1}^{n} (\mu_{kj})^q DIS_{kj}^2         
	\end{equation}
	subject to:
	\begin{equation}
	0 \leq \mu_{kj}\leq1;
	\end{equation}
	\begin{equation}
	\sum_{k=1}^{c} \mu_{kj}=1
	\end{equation}
	\begin{equation}
	0<\sum_{j=1}^{n} \mu_{kj} < n; 
	\end{equation}
	Where: \\
	
		\noindent $n$= number of data\\
		$c$= number of clusters (topics)\\
		$\mu_{kj}$= membership value\\
		$q$= fuzzifier, $1 < q \le \infty$ \\
		$V$= cluster center vector\\
		$DIS_{kj}=d(x_j,v_k)$= distance between $x_j$ and $v_k$ \\

	\noindent By optimizing eq.3:
	\begin{equation}
	\mu_{ij}= \frac{1}{\sum_{l=1}^{c} (\frac{DIS_{kj}}{DIS_{lj}})^ \frac{2}{q-1}}
	\end{equation}
	\begin{equation}
	v_i=\frac{\sum_{j=1}^{n} (\mu_{kj})^q x_j}{\sum_{j=1}^{n} (\mu_{kj})^q}
	\end{equation}
	
	The iterations in the clustering algorithm continue till the maximum changes in $\mu_{kj}$ becomes less than or equal to a pre-specified threshold with $O(n)$ computational time complexity.  
	
	We use $\mu_{kj}$ as the membership degrees for each document (D) with respect to each of topics (clusters). The value of $\mu_{kj}$ is between 0 and 1 that can be interpreted as $\mathbf{\underline{P(T_k|D_j)}}$ or probability of topic k in document j. This step finds $Topics \times Documents$ matrix  and uses it along with the following steps to find $Words \times Topics$ matrix in Figure \ref{tab: drtm}. It is worth mentioning that FLAS's steps are dependent and integrated. For example, the documents (D) and the topics (T) in step 3 are the same topics and documents in steps 5,6 and 7. 
	
	To avoid the negative impact of high dimensionality of the four mentioned matrices in step 2, we use singular value decomposition (SVD), which is a popular method \cite{fodor2002survey} to reduce the data dimension before using fuzzy clustering. We select two dimensions, as the minimum number of dimensions, for SVD to have a fast process. \\
	
	\textbf{Step 4.} 
	We use document-term matrices with the GTW methods in step 2 ($Words \times Documents$ matrix) to find $\mathbf{\underline{P(D_j})}$ or probability of document j:
	
	\begin{equation}
	P(D_j)=\frac{\sum_{i=1}^{m} (W_i,D_j)}{\sum_{i=1}^{m}\sum_{j=1}^{n} (W_i,D_j)}
	\end{equation}

	\textbf{Step 5.} The next step has two parts. The first part is to find $\mathbf{\underline{P(D_j|T_k)}}$ or probability of document j in topic k  using $P(T_k|D_j)$ in step 3 and $P(D_j)$ in step 4: 
	\begin{equation}
	P(D_j,T_k)=P(T_k|D_j) \times P(D_j)
	\end{equation}
	Then normalizing $P(D,T)$ in each topic:
	\begin{equation}
	P(D_j|T_k)=\frac{P(D_j,T_k)}{\sum_{j=1}^{n} P(D_j,T_k)}
	\end{equation}
	\textbf{Step 6:} We use document-term matrices with the GTW methods in step 2 to find $\mathbf{\underline{P(W_i|D_j)}}$ or probability of word i in document j:
	\begin{equation}
	P(W_i|D_j)=\frac{P(W_i,D_j)}{\sum_{i=1}^{m} P(W_i,D_j)}
	\end{equation}
	\textbf{Step 7:} The final step is to find \underline{$\mathbf{P(W_i|T_k)}$} or probability of word i in topic k ($Topics \times Documents$  matrix in Figure \ref{tab: drtm}) using $P(D_j|T_k)$ in step 5 and $P(W_i|D_j)$ in step 6:
	\begin{equation}
	P(W_i|T_k)=\sum_{j=1}^{n} P(W_i|D_j) \times P(D_j|T_k)
	\end{equation}

	FLSA is flexible to work with all dimensionality reduction techniques \cite{fodor2002survey} such as principal component analysis (PCA) \cite{jolliffe2002principal} and all fuzzy clustering techniques such as self-organizing map (SOM) \cite{kohonen1990self}. 
	
	Fuzzy c-means, as it is the core of FLSA, can be applied on both discrete and continuous data. This feature enables FLSA to use a wide range of LTW and GTW methods and to be used for other machine learning and data science applications such as image processing \cite{keller2005fuzzy}. Moreover, optimization nature of fuzzy clustering provides a solution to estimate the optimum number of topics.

	\section{Experiments}
	\label{Ex}
	
	In this section, we evaluate FLSA against LDA by document classification using Random Forest, document clustering using k-means, document modeling using log-likelihood, and execution time test. We also evaluate FLSA against RedLDA by document modeling on redundant documents. 
	
	We use five datasets, the Matlab package for Chib-style estimation of log-likelihood\footnote{\url{http://www.cs.umass.edu/~wallach/code/etm/}},  the FCM Matlab package\footnote{\url{http://www.mathworks.com/help/fuzzy/fcm.html}} with its default settings including 100 iterations and 1e-5 as the minimum improvement in objective function between two consecutive iterations, the Weka tool\footnote{\url{http://www.cs.waikato.ac.nz/ml/weka/}} for classification evaluation, the MALLET package\footnote{\url{http://mallet.cs.umass.edu/}} with its default settings for implementing LDA, and the Python package for implementing RedLDA\footnote{\url{https://sourceforge.net/projects/redlda/}}. The source code for FLSA will be available in the first author's website\footnote{\url{https://sites.google.com/site/karamihomepage/}}$^{,}$\footnote{\url{https://github.com/amir-karami}} in R and Matlab platforms.

		\begin{table*}[ht]
			\normalsize
			\centering
			\scalebox{0.7}{
				\begin{tabular}{|p{2.7cm}|p{3cm}|p{3cm}|p{3.5cm}|p{4.5cm}|p{2.6cm}|} \hline
					\textbf{Dataset Name} & \textbf{\#Documents} & \textbf{\#Term Tokens} & \textbf{\#Unique Terms} & \textbf{Avg Term Per Document} & \textbf{Description} \\ \hline
					\textbf{M-Dataset} & 1,527 & 245,931 & 14,411 & 96.3 &  Medical Papers \\ \hline
					\textbf{N-Dataset} & 1,607 & 299,449 & 11,059  & 124.8 & Nursing Notes\\ \hline
					\textbf{O-Dataset} & 2,092 & 198,998 & 15,768 &  95.1 & Medical Papers\\ \hline
					\textbf{T-Dataset} & 58,927 & 395,635 &  25,310 & 6.7 & Tweets\\ \hline
					\textbf{R-Dataset-1} & 2,288	&  754,801 & 3,265 & 231.2 &  News\\ \hline
					\textbf{R-Dataset-2}& 3,310 & 785,467 & 23,408 &  236.4 & News\\ \hline
					\textbf{R-Dataset-3}& 3,254 & 757,603 & 22,982 & 232.8 &  News\\ \hline
					\textbf{R-Dataset-4}& 3,162 & 705,100 & 22,264 & 222.9 & News\\ \hline
					\textbf{R-Dataset-5}&  3,211& 751,496 & 23,046 & 234.1 &  News\\ \hline
					\textbf{R-Dataset-6}& 3,124 & 727,687  & 22,944 & 232.9 & News\\ \hline
					\textbf{R-Dataset-7}& 3,251 & 747,067  & 22,609 &  229.8 & News \\ \hline
					\textbf{R-Dataset-8}& 3,212  & 747,956 & 22,961 &  232.6 & News \\ \hline
					\textbf{R-Dataset-9} &  3,257& 755,261 & 22,929 &  231.9 & News \\ \hline
					\textbf{R-Dataset-10} & 3,258 & 739,041 & 22,913 &  226.8 & News \\ \hline
					\textbf{R-Dataset-11}  &  3,284 & 764,658 & 23,134 &  232.8 & News \\ \hline  
					
				\end{tabular}}
				
				\caption{Basic Statistics for Datasets}
				\label{tab: Datasets}
			\end{table*}		
		
		\vspace{-5mm}
		\subsection{Datasets}
		We leverage five available health and medical datasets in this research (Table~\ref{tab: Datasets}):
		
		\begin{itemize}
			\itemsep0em 
			
			\item The first  dataset\footnote{\url{http://muchmore.dfki.de/resources1.htm}} is MuchMore Springer Bilingual Corpus (M-Dataset) which is a labeled corpus of English scientific medical abstracts from the Springer website. In this research, we use the first 2 journals including: Arthroscopy and Federal Health Standard Sheet.
			
			\item The second dataset\footnote{\url{http://physionet.org/}} is an unlabeled corpus of 2,434 nursing notes (N-Dataset).

			\item The third  dataset\footnote{\url{http://disi.unitn.it/moschitti/corpora/ohsumed-first-20000-docs.tar.gz}} is Ohsumed Collection (O-Dataset) that is a labeled corpus of medical abstracts from the MeSH categories including Bacterial Infections and Mycoses, and Virus Diseases.

			\item The fourth dataset (T-Dataset) is health news from Twitter\footnote{\url{www.twitter.com}}. We collected the tweets from the health related Twitter accounts including \textit{cbchealth, everydayhealth, foxnewshealth, goodhealth, kaiserhealthnews, latimeshealth, msnhealthnews, NBChealth, nprhealth, usnewshealth, bbchealth, cnnhealth, gdnhealthcare, nytimeshealth, reuters\_health,} and \textit{wsjhealth} from August 2011 to December 2014\footnote{\url{https://github.com/amir-karami/Health-News-Tweets-Data}}. 
			
			\item The fifth dataset is R-Datasets\footnote{\url{https://sourceforge.net/projects/corpusredundanc/files/?source=navbar}} are the synthesis text documents to track the negative effect of redundancy in documents \cite{cohen2013redundancy}. These datasets are subsets of a larger dataset called \textit{WSJ} which has a collection of the abstracts of Wall Street Journal. In this dataset, 1300 abstracts were sampled between 1 and 5 times in a uniform manner for 11 times to eliminate bias from random sampling. 
		\end{itemize}

	\begin{table*}[ht]
		\centering
\large
		
		\scalebox{0.6}{
			\begin{tabular}{|p{3.5cm}|p{1.2cm}|p{2.5cm}|p{1cm}|p{1cm}|p{1.7cm}|p{1.8cm}|}
				\hline
				
				\textbf{Method}  & \textbf{Acc \%} & \textbf{F-Measure} &  \textbf{MCC}&\textbf{ROC} &  \textbf{\#Topics}  \\ \hline
				
				LDA &  90.05  & 0.9  & 0.799 & 0.969 &  50 \\ \hline
				
				\textbf{FLSA(Entropy)}&  \textbf{97.66} & \textbf{0.977} & \textbf{0.953} & \textbf{0.99}  & 50\\ \hline

				FLSA(IDF) &   95.90 & 0.959 & 0.917 & 0.982 & 50\\ \hline
				
				FLSA(Normal)  &  91.22 & 0.912 & 0.824 & 0.971 &  50\\ \hline

				FLSA(ProbIDF) &  97.66 & 0.977 & 0.953 & 0.987  & 50\\ \hline
				
			\end{tabular}
			\qquad
			\begin{tabular}{|p{3.5cm}|p{1.2cm}|p{2.5cm}|p{1cm}|p{1cm}|p{1.7cm}|p{1.8cm}|}
				\hline
				
				\textbf{Method}  & \textbf{Acc \%} & \textbf{F-Measure} &  \textbf{MCC}&\textbf{ROC} &  \textbf{\#Topics}  \\ \hline
				
				LDA &  78.36  & 0.78  & 0.561 & 0.895 &  100 \\ \hline
				
				FLSA(Entropy)&  96.49 & 0.965 & 0.929 & 0.986 &  100\\ \hline

				\textbf{FLSA(IDF)} &   \textbf{98.24} & \textbf{0.982} & \textbf{0.964} & \textbf{0.996} &  100\\ \hline
				
				FLSA(Normal) &  92.39 & 0.924 & 0.846 & 0.984 &  100\\ \hline

			FLSA(ProbIDF) &  97.66 & 0.977 & 0.953 & 0.994 &  100\\ \hline
				
			\end{tabular}} 
			
			\caption{M-Datatset Classification - 50 and 100 Topics}
			\label{tab: 1-2-50-100}
			
			\vspace{3mm}
			
			\scalebox{0.6}{
				\begin{tabular}{|p{3.5cm}|p{1.2cm}|p{2.5cm}|p{1cm}|p{1cm}|p{1.7cm}|p{1.8cm}|}
					\hline
					
					\textbf{Method}  & \textbf{Acc \%} & \textbf{F-Measure} &  \textbf{MCC}&\textbf{ROC} &  \textbf{\#Topics}  \\ \hline
					
					LDA &  77.19  & 0.77  & 0.536 & 0.867 &  150 \\ \hline
					
					FLSA(Entropy)&  95.90 & 0.959 & 0.917 & 0.995 &  150\\ \hline

					FLSA(IDF) &   97.66 & 0.977 & 0.953 & 0.991 &  150\\ \hline
					
					FLSA(Normal)  &  95.32 & 0.953 & 0.905 & 0.992 &  150\\ \hline

					\textbf{FLSA(ProbIDF)} &  \textbf{97.07} & \textbf{0.971} & \textbf{0.941} & \textbf{0.991} &  150\\ \hline
					
				\end{tabular}
				\qquad
				\begin{tabular}{|p{3.5cm}|p{1.2cm}|p{2.5cm}|p{1cm}|p{1cm}|p{1.7cm}|p{1.8cm}|}
					\hline
					
					\textbf{Method}  & \textbf{Acc \%} & \textbf{F-Measure} &  \textbf{MCC}&\textbf{ROC} &  \textbf{\#Topics}  \\ \hline
					
					LDA &  82.45  & 0.822  & 0.646 & 0.894 &  200 \\ \hline
					
					FLSA(Entropy) &  97.076 & 0.971 & 0.941 & 0.992  & 200\\ \hline

				FLSA(IDF) &   97.66 & 0.977 & 0.953 & 0.984 & 200\\ \hline
					
					FLSA(Normal)  &  92.39 & 0.924 & 0.846 & 0.982 &  200\\ \hline

					\textbf{FLSA(ProbIDF)} &  \textbf{97.66} & \textbf{0.977} & \textbf{0.953} & \textbf{0.985} &  200\\ \hline
					
				\end{tabular}
			} 
			
			\caption{M-Datatset Classification - 150 and 200 Topics}
			
			\label{tab: 1-2-150-200}
			
			\vspace{3mm}
			
			\scalebox{0.6}{
				\begin{tabular}{|p{3.5cm}|p{1.2cm}|p{2.5cm}|p{1cm}|p{1cm}|p{1.7cm}|p{1.8cm}|}
					\hline
					
					\textbf{Method}  & \textbf{Acc \%} & \textbf{F-Measure} &  \textbf{MCC}&\textbf{ROC} &  \textbf{\#Topics}  \\ \hline
					
					LDA &  75.38  & 0.72  & 0.282 & 0.66 &  50 \\ \hline
					
					FLSA(Entropy)&  75.21 & 0.741 & 0.331 & 0.737 &  50\\ \hline

					\textbf{FLSA(IDF)} &   \textbf{75.90} & \textbf{0.746} & \textbf{0.343} & \textbf{0.727} &  50\\ \hline
					
					FLSA(Normal)  &  71.25 & 0.677 & 0.153 & 0.63 &  50\\ \hline

					FLSA(ProbIDF) &  74.87 & 0.735 & 0.314 & 0.711 &  50\\ \hline
					
				\end{tabular}
				\qquad
				\begin{tabular}{|p{3.5cm}|p{1.2cm}|p{2.5cm}|p{1cm}|p{1cm}|p{1.7cm}|p{1.8cm}|}
					\hline
					
					\textbf{Method}  & \textbf{Acc \%} & \textbf{F-Measure} &  \textbf{MCC}&\textbf{ROC} &  \textbf{\#Topics}  \\ \hline
					
				LDA &  72.97  & 0.682  & 0.179 & 0.657 &  100 \\ \hline
					
					\textbf{FLSA(Entropy)}&  \textbf{76.24} & \textbf{0.747} & \textbf{0.344} & \textbf{0.732}  & 100\\ \hline

					FLSA(IDF) &   74.35 & 0.726 & 0.288 & 0.712  & 100\\ \hline
					
					FLSA(Normal)  &  71.08 & 0.694 & 0.201 & 0.617  & 100\\ \hline

					FLSA(ProbIDF) & 74.52 & 0.724 & 0.283 & 0.684 &  100\\ \hline
					
				\end{tabular}
			} 
			
			\caption{O-Datatset Classification- 50 and 100 Topics}
			\label{tab: 2-2-50-100}
			\vspace{3mm}

			\scalebox{0.6}{
				\begin{tabular}{|p{3.5cm}|p{1.2cm}|p{2.5cm}|p{1cm}|p{1cm}|p{1.7cm}|p{1.8cm}|}
					\hline
					
					\textbf{Method}  & \textbf{Acc \%} & \textbf{F-Measure} &  \textbf{MCC}&\textbf{ROC} &  \textbf{\#Topics}  \\ \hline
					
					LDA &  72.80  & 0.662  & 0.136 & 0.636 &  150 \\ \hline
					
					FLSA(Entropy)&  74.87 & 0.735 & 0.312 & 0.723 &  150\\ \hline

					\textbf{FLSA(IDF)} &   \textbf{76.59} & \textbf{0.752} & \textbf{0.358} & \textbf{0.732} &  150\\ \hline
					
				FLSA(Normal)  &   72.46 & 0.691 & 0.194 & 0.668 &  150  \\ \hline

					FLSA(ProbIDF) & 75.04 & 0.735 & 0.313 & 0.726 &  150\\ \hline
					
				\end{tabular}
				\qquad
				\begin{tabular}{|p{3.5cm}|p{1.2cm}|p{2.5cm}|p{1cm}|p{1cm}|p{1.7cm}|p{1.8cm}|}
					\hline
					
					\textbf{Method}  & \textbf{Acc \%} & \textbf{F-Measure} &  \textbf{MCC}&\textbf{ROC} &  \textbf{\#Topics}  \\ \hline
					
					LDA &  71.08  & 0.648  & 0.079 & 0.63 &  200 \\ \hline
					
					\textbf{FLSA(Entropy)}&  \textbf{75.21} & \textbf{0.74} & \textbf{0.326} & \textbf{0.731} &  200\\ \hline

					FLSA(IDF) &   74.18 & 0.725 & 0.285 & 0.713 &  200\\ \hline
					
					FLSA(Normal)  &  71.94 & 0.683 & 0.172 & 0.657 &  200\\ \hline

					FLSA(ProbIDF) &  74.87 & 0.729 & 0.298 & 0.719 &  200\\ \hline
					
				\end{tabular}
			} 
			
			\caption{O-Datatset Classification - 150 and 200 Topics}
			\label{tab: 2-2-150-200}
			
		\end{table*}

					\vspace{-5mm}
			\subsection{Document Classification}
			
			The first evaluation measure is document classification using two labeled datasets, M-Dataset and O-Dataset. Both datasets have two classes (labels), M-Dataset with Arthroscopy and Federal Health Standard Sheet classes, and O-Dataset with Bacterial Infections and Mycoses, and Virus Diseases classes.
			
			Document classification problem assigns a document to a class and this problem needs to extract features from text data. To avoid high dimensionality of BOW approach for document classification, topic modeling reduces the number of features by clustering the meaningful related words as a topic. To avoid any possible bias, we track the performance of FLSA against LDA with the 10-fold cross validation method that the data is broken into 10 subsets for 10 iterations. Each of the subsets is selected for testing and the rest of sets are selected for training. We use 50, 100, 150, and 200 topics as the input features of documents for Random Forest method as it is one of the popular and high performance classification methods \cite{qi2006evaluation}. The output of Random Forest is presented as a confusion matrix (Table~\ref{tab: confmx}) with the following definitions:
			
			\begin{itemize}
				\item True Negative (TN) is the number of correct predictions that an instance is negative.
				
				\item False Negative (FN) is the number of incorrect of predictions that an instance negative.
				
				\item False Positive (FP) is the number of incorrect predictions that an instance is positive.
				
				\item True Positive (TP) is the number of correct predictions that an instance is positive.
			\end{itemize}
			
			\begin{table}[H]
				\centering
				
				\begin{tabular}{cc|c|c|c|}
					\cline{3-4}
					& & \multicolumn{2}{ c| }{\textbf{Predicted}} \\ \cline{3-4}
					& & \textbf{Negative} & \textbf{Positive}  \\ \cline{1-4}
					\multicolumn{1}{ |c| }{\multirow{2}{*}{\textbf{Actual}} } &
					\multicolumn{1}{ |c| }{\textbf{Negative}} & TN & FP    \\ \cline{2-4}
					\multicolumn{1}{ |c  }{}                        &
					\multicolumn{1}{ |c| }{\textbf{Positive}} & FN & TP      \\ \cline{1-4}
					
				\end{tabular}

				\caption{Confusion Matrix}
				\label{tab: confmx}
			\end{table}

			For  evaluating the performance of the classification algorithm, we use accuracy (ACC), F-measure, Matthews Correlation Coefficient (MCC), and the area under ROC (AUC). The evaluation metrics are defined based on the confusion matrix, as shown in equations~\ref{eq: prec}~-~\ref{eq: mcc}: 
			
\begin{figure*}[htp!]
	\centering
	\normalsize
	
	\begin{tikzpicture}[scale=.6]
	\begin{axis}
	[xlabel=\# Clusters,ylabel= Calinski-Harabasz Index,legend style={at={(0.5,-0.2)}, anchor=north,legend columns=-1}]
	\addplot coordinates { (2,5234) (3,6473) (4,7855) (5,8528) (6,10193) (7,10636)(8,13873)};
	\addplot coordinates { (2,5105) (3,6332) (4,7554) (5,8026) (6,10024) (7,9996)(8,13411)};
	\addplot coordinates { (2,5022) (3,6068) (4,7230) (5,8469) (6,9407) (7,11196)(8,12345)};
	\addplot coordinates { (2,5520) (3,6677) (4,7079) (5,10148) (6,11423) (7,15153)(8,15710)};
	\addplot coordinates { (2,174) (3,140) (4,138) (5,133) (6,130) (7,117)(8,119)};
	\end{axis}
	\end{tikzpicture}
	\qquad
	\begin{tikzpicture}[scale=.6]
	\begin{axis}
	[xlabel=\# Clusters,ylabel= Calinski-Harabasz Index,legend style={at={(0.5,-0.2)}, anchor=north,legend columns=-1}]
	\addplot coordinates { (2,5234) (3,6473) (4,7855) (5,8528) (6,10193) (7,10636)(8,13873)};
	\addplot coordinates { (2,5105) (3,6332) (4,7554) (5,8026) (6,10024) (7,9996)(8,13411)};
	\addplot coordinates { (2,5022) (3,6068) (4,7230) (5,8469) (6,9407) (7,11196)(8,12345)};
	\addplot coordinates { (2,5510) (3,6666) (4,7067) (5,10127) (6,11398) (7,15110)(8,15922)};
	\addplot coordinates { (2,115) (3,90) (4,83) (5,76) (6,82) (7,75)(8,77)};
	\end{axis}
	
	\end{tikzpicture}
	
	\begin{tikzpicture}

	\begin{customlegend}[legend columns=-1,
	legend style={
		draw=none,
		column sep=1ex,font=\small
	},legend entries={$FLSA(Entropy)$,$FLSA(IDF)$,$FLSA(ProbIDF)$,$FLSA(Normal)$,$LDA$}]
	
	\addlegendimage{blue,fill=blue,mark=*,sharp plot}
	\addlegendimage{red,fill=red!50!black,mark=square*,sharp plot}
	\addlegendimage{brown,fill=black!50!brown,mark=otimes*,sharp plot}
	\addlegendimage{black,fill=black,mark=star,sharp plot}
	\addlegendimage{blue,fill=blue,mark=diamond*,sharp plot}
	\end{customlegend}
	\end{tikzpicture}
	
	\caption{Calinski-Harabasz - 50 Topics Left Figure and 100 Topics Right Figure}
	\label{fig:ch-50-100}
	
	\vspace{2mm}
	
	\begin{tikzpicture}[scale=.6]
	\begin{axis}
	[xlabel=\# Clusters,ylabel= Calinski-Harabasz Index,legend style={at={(0.5,-0.2)}, anchor=north,legend columns=-1}]
	\addplot coordinates { (2,5232) (3,6470) (4,7854) (5,8528) (6,10189) (7,10632)(8,13870)};
	\addplot coordinates { (2,5105) (3,6332) (4,7554) (5,8026) (6,10024) (7,9996)(8,13411)};
	\addplot coordinates { (2,5022) (3,6068) (4,7230) (5,8469) (6,9407) (7,11196)(8,12345)};
	\addplot coordinates { (2,5507) (3,6663) (4,7062) (5,10120) (6,11391) (7,15106)(8,15914)};
	\addplot coordinates { (2,63) (3,59) (4,57) (5,57) (6,54) (7,49)(8,53)};
	\end{axis}
	
	\end{tikzpicture}
	\qquad
	\begin{tikzpicture}[scale=.6]
	\begin{axis}
	[xlabel=\# Clusters,ylabel= Calinski-Harabasz Index,legend style={at={(0.5,-0.2)}, anchor=north,legend columns=-1}]
	\addplot coordinates { (2,5232) (3,6470) (4,7854) (5,8528) (6,10189) (7,10632)(8,13870)};
	\addplot coordinates { (2,5105) (3,6332) (4,7554) (5,8026) (6,10024) (7,9996)(8,13411)};
	\addplot coordinates { (2,5022) (3,6068) (4,7230) (5,8469) (6,9407) (7,11196)(8,12345)};
	\addplot coordinates { (2,5507) (3,6663) (4,7062) (5,10120) (6,11391) (7,15106)(8,15914)};
	\addplot coordinates { (2,63) (3,59) (4,57) (5,57) (6,54) (7,49)(8,53)};
	
	\end{axis}

	\end{tikzpicture}
	\begin{tikzpicture}

	\begin{customlegend}[legend columns=-1,
	legend style={
		draw=none,
		column sep=1ex,font=\small
	},legend entries={$FLSA(Entropy)$,$FLSA(IDF)$,$FLSA(ProbIDF)$,$FLSA(Normal)$,$LDA$}]
	
	\addlegendimage{blue,fill=blue,mark=*,sharp plot}
	\addlegendimage{red,fill=red!50!black,mark=square*,sharp plot}
	\addlegendimage{brown,fill=black!50!brown,mark=otimes*,sharp plot}
	\addlegendimage{black,fill=black,mark=star,sharp plot}
	\addlegendimage{blue,fill=blue,mark=diamond*,sharp plot}
	\end{customlegend}
	\end{tikzpicture}
	
	\caption{Calinski-Harabasz - 150 Topics Left Figure and 200 Topics Right Figure}
	\label{fig:ch-150-200}
\end{figure*}

			\begin{equation} 
			Precision(P)= \frac{TP}{TP+FP}
			\label{eq: prec}
			\end{equation}
			\begin{equation} 
			Recall(R)= \frac{TP}{TP+FN}
			\end{equation}
			\begin{equation} 
			Accuracy (Acc)= \frac{TP+TN}{TP+TN+FP+FN}
			\end{equation}
			\begin{equation} 
			F-measure= 2\times\frac{P \times R}{P+R}
			\label{eq: fmeasure}
			\end{equation}
			
			ROC curves plot FP on the X axis vs. TP on the Y axis to find the trade off between them; therefore, the ROC is closer to the upper left indicating better performance (Figure~\ref{fig:roc}). 
			
			\begin{figure}[ht]
				\centering
				\scalebox{1}{
					\begin{tikzpicture}[scale=2]

					\draw [help lines] (0,0) grid (1,1);
					\draw[thick,->] (0,0) -- (1,0) node[anchor=north west] {FP};
					\draw[thick,->] (0,0) -- (0,1)node[anchor=south east] {TP};
					\draw[color=red] (0,0) to (1,1);
					\draw [color=blue](0,0) to[out=90,in=180] (1,1);
					
					\end{tikzpicture}}
				\caption{ROC}
				\label{fig:roc}
			\end{figure}

			MCC is used to determine the quality of classification methods, ranging between -1 (the worst performance) and +1 (the best performance). \\
			
MCC=
			\begin{equation}
			\resizebox{0.48\textwidth}{!}{
			$\frac{(TP \times TN)-(FN \times FP)}{\sqrt{(TP+FP)\times (TP+FN)\times (TN+FP)\times(TN+FN)}}$}
					\label{eq: mcc}
			\end{equation}

	 This experiment shows that FLSA with Entropy, IDF, Normal, and ProbIDF show better performance than LDA with different numbers of topics (Tables~\ref{tab: 1-2-50-100}~-~\ref{tab: 2-2-150-200}). The highest performance in each table is shown in bold format.

\subsection{Document Clustering}
\label{ssec:layout}
The second evaluation is document clustering using unlabeled N-Dataset. Internal and external validation are two major methods for clustering validation; however, comparison between these two major methods shows that internal validation is more precise \cite{rendon2011internal}. We evaluate different numbers of topics and clusters with Calinski-Harabasz (CH) index, as one of the popular internal validation methods, using K-means with 500 iterations. CH index evaluates the cluster validity based on the average of the sum of squared error cluster between and within clusters. Higher CH index indicates “better” clustering. \\ 
\indent We track the performance of FLSAs and LDA using different numbers of clusters ranging from 2 to 8 with different numbers of topics including 50, 100, 150, and 200. CH index shows that FLSAs have better performance than LDA with the different ranges of features and clusters (Figures~\ref{fig:ch-50-100} \& \ref{fig:ch-150-200}). The gap between FLSAs and LDA does not change significantly with different numbers of topics and clusters.

			\vspace{-5mm}			
\subsection{Redundancy Issue}
The next experiment explores the effect of redundancy issue using the fifth dataset. R-Datasets are not a medical corpus; however, they were created as a synthetic redundant corpus without having privacy issue to measure the effect of redundancy issue \cite{cohen2013redundancy}. We select publicly available unlabeled R-Datasets to make the evaluation process easier for possible future research. We compare FLSAs with not only LDA but also RedLDA, as it was developed to handle redundancy issue in medical text data \cite{cohen2014redundancy}. 

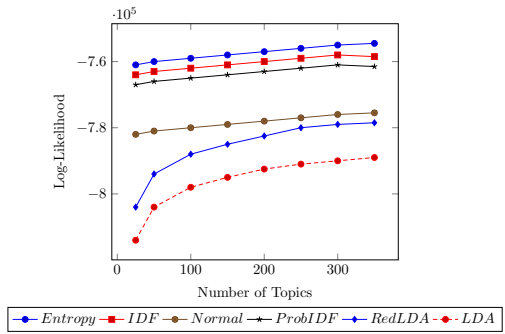
\begin{figure}[H]
			\normalsize
			\centering
			
			\begin{tikzpicture}[scale=0.55]
			\begin{axis}[ xlabel={Number of Topics}, ylabel={Log-Likelihood},legend style={at={(0.5,-0.2)}, anchor=north,legend columns=-1}]
			\addplot coordinates { (25,-761000)(50,-760000)(100,-759000)(150,-758000)(200,-757000) (250,-756000) (300,-755000)(350,-754500)};
			
			\addplot coordinates { (25,-764000)(50,-763000)(100,-762000)(150,-761000)(200,-760000) (250,-759000) (300,-758000)(350,-758500)};
			
			\addplot coordinates { (25,-782000)(50,-781000)(100,-780000)(150,-779000)(200,-778000) (250,-777000) (300,-776000)(350,-775500)};
			
			\addplot coordinates { (25,-767000)(50,-766000)(100,-765000)(150,-764000)(200,-763000) (250,-762000) (300,-761000)(350,-761500)};
			
			\addplot coordinates { (25,-804000)(50,-794000)(100,-788000)(150,-785000)(200,-782500) (250,-780000) (300,-779000)(350,-778500)};
			
			\addplot coordinates { (25,-814000)(50,-804000)(100,-798000)(150,-795000)(200,-792500) (250,-791000) (300,-790000)(350,-789000) };
			\legend{$Entropy$, $IDF$,$Normal$, $ProbIDF$, $RedLDA$,$LDA$}
			\end{axis}
			\end{tikzpicture}
			
			\caption{Likelihood Comparison for R-Datasets}
			\label{tab: log-R}

		\end{figure}

We train LDA, RedLDA and FLSA models on R-Datasets to compare the generalization performance of the models. We compute the log-likelihood on a held-out test set to evaluate the models. A higher log-likelihood score indicates better generalization performance. Figure~\ref{tab: log-R} shows the average log-likelihood of the R-Datasets with different numbers of topics from 25 to 350. This experiment indicates that FLSAs performs better than RedLDA and LDA on redundant documents.

\subsection{Execution Time}

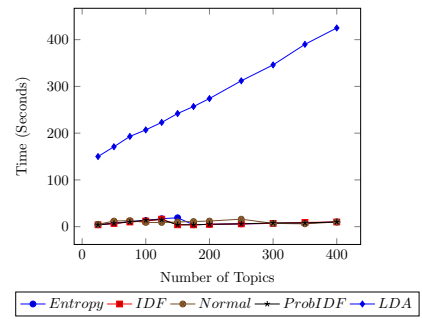
\begin{figure}[H]
					\centering
								\normalsize
					\begin{tikzpicture}[scale=0.55]
					\begin{axis}[ xlabel={Number of Topics}, ylabel={Time (Seconds)},legend style={at={(0.5,-0.2)}, anchor=north,legend columns=-1}]
					\addplot coordinates { (25,4)(50,7)(75,11) (100,13)(125,17) (150,19)(175,4) (200,5) (250,6) (300,7)(350,8) (400,10)};
					\addplot coordinates { (25,4)(50,7)(75,10) (100,12)(125,16) (150,4)(175,4) (200,5) (250,6) (300,7)(350,9) (400,10) };
					\addplot coordinates { (25,5)(50,12)(75,13) (100,9)(125,9) (150,10)(175,11) (200,12) (250,16) (300,7)(350,6) (400,10) };
					\addplot coordinates { (25,4)(50,7)(75,10) (100,14)(125,15) (150,4)(175,4) (200,5) (250,6) (300,8)(350,8) (400,10) };
					\addplot coordinates { (25,150)(50,171)(75,193) (100,207)(125,223) (150,242)(175,257) (200,274) (250,312) (300,346)(350,390) (400,425)};
					\legend{$Entropy$, $IDF$,$Normal$, $ProbIDF$, $LDA$}
					\end{axis}
					\end{tikzpicture}

					\caption{Execution Time for T-Dataset}
					\label{fig: T-dataset}
										\vspace{-3mm}
				\end{figure}

In this section, we compare the speed of FLSA in comparison with LDA using T-Dataset, the biggest dataset in this paper. The major process in topic modeling is based on a joint probability distribution over hidden topics and the observed words to infer the word with higher probability in each topic by using the posterior distribution. The most popular approximate method for LDA is collapsed Gibbs sampling applied in the experiments. All the inference algorithms require multiple iterations which increase the computational cost linearly with the number of documents, topics, words, and iterations \cite{zeng2012new}. Figure~\ref{fig: T-dataset} shows that the time performance of FLSAs is stable with an increase in the numbers of topics and considerably better than LDA.

	\section{Example}

\begin{figure}[ht]
					\centering
								\normalsize
					\begin{tikzpicture}[scale=0.55]
					\begin{axis}[ xlabel={Number of Topics}, ylabel={Objective Function (J)},legend style={at={(0.5,-0.2)}, anchor=north,legend columns=-1}]
					\addplot coordinates { (25,0.0561)(50,0.0307)(75,0.0211) (100,0.0165)(125,0.0134) (150,0.0114)(175,0.0111) (200,0.0097) (250,0.0078) (300,0.0065) (350,0.0056) (400,0.0049)};
					
					\end{axis}
					\end{tikzpicture}
					\caption{Number of Topics Estimation}
					\label{fig: numTopics}
				\end{figure}
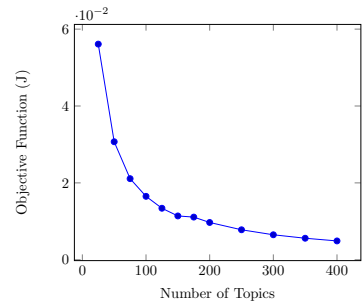

Documents and papers in health and medical domains contain terms and words that need expertise to understand them. Therefore, we show an illustrative example with T-Dataset containing tweets with more understandable words in this section. First, we run FLSA on different numbers of topics to estimate optimum number of topics using the objective function ,$J_q$ (Figure~\ref{fig: numTopics}). This estimation feature of fuzzy clustering is on of advantageous of FLSA. The elbow or the knee part of the figure shows $\sim$125 as the number of topics\footnote{\url{https://www.cs.princeton.edu/courses/archive/spring07/cos424/scribe_notes/0306.pdf}}.

Table~\ref{tab: cluster3} shows a sample of topics that are about family, cancer, Ebola, Alzheimer, nursing, and teenagers:

\begin{itemize}
\item $T_{83}$ is about the benefit of nursing black teens with dementia.
				
\item $T_{42}$ is about the huge costs of teens pregnancy.

\item $T_{33}$  is reporting a teen's invention to save Alzheimer's patients and improve the quality of their lives.
		
\item $T_{101}$ is about the government's need to hire more nurses in different cities.

\item $T_{85}$ is about the recipes for family having diabetes specially for women and kids.
				
\item $T_{107}$ is about using Ebola for cyberbullying.
				
\item $T_{71}$  is about the role of psychiatrists against Ebola.
				
\item $T_{25}$ is about sneezing as a risk for spreading Ebola and the relation between Ebola and lung cancer. 
							
\end{itemize}

\begin{table*}[ht]
	\centering
	\Large
	\scalebox{0.6}{
		\begin{tabular}{ |p{3cm}|p{3cm}|p{3cm}|p{3cm}|p{3cm}|p{3cm}|p{3cm}|p{3cm}| }
			\hline
			
			$T_{83}$   & $T_{42}$  & $T_{33}$   & $T_{101}$ & $T_{85}$ & $T_{107}$  & $T_{71}$  & $T_{25}  $     \\
			\hline
			teen         & sick     & teens     & deadly  & diabetes &  rt            & rt            & rt      \\
			nursing      & pregnant & put       & government					     & week     &   health        & health        & health        \\
			dementia     & billion  & popular   & nurses  					     & find     &   ebola         & ebola         & ebola        \\
			benefit       & clinical & save      & affordable					    & recipes  &     lightning     & pms           & study         \\
			ready       & dying    & alzheimer & clinic 					    & work     &     study         & poached       & cancer     \\
			babies        & stories  & leave     & premiums					   & low      &      welcomes      & upset         & care      \\
			race         & rare     & run       & ban  					    & kids     &  vape          & study         & amp\\
			black        & access   & docs      & city 					  & rt       &    cyberbullying & tension       & sneeze         \\
			worst     & role     & quality   & warns  					 & women    &      visually      & psychiatrists & poo       \\
			meet        & hope     & tied      & deadline 					    & life     &   tremors       & embryo        & risk       \\
			\hline

		\end{tabular}}
		
		\caption{A Sample of Topics}
		\label{tab: cluster3}
	\end{table*}

\section{Conclusion}
\label{Co}

The vast array of health and medical text data represents a valuable resource that can be analyzed to advance state-of-the-art medicine and health. Large electronic health and medical archives such as PubMed provide an extremely useful service to the scholarly community. However, the needs of readers go beyond a simple keyword search.

Topic modeling is one of the popular unsupervised methods to automatically discover a hidden thematic structure in a large collection of unstructured health and medical documents. This discovered structure facilitates browsing, searching, and summarizing the collection. 
				
Fuzzy perspective is a machine learning approach that has been used more in medical image processing than text processing. Existing techniques of topic modeling are based on two main approaches: linear algebra and statistical distributions; however, this paper proposes FLSA to utilize fuzzy perspective for disclosing latent semantic features of health and medical text data.

FLSA is a new competitor to the established topic models such as LDA and has the flexibility to work with a wide range of dimension reduction and fuzzy clustering techniques. FLSA also works with both discrete and continuous data, estimates the optimum number of topics, and avoids the negative effect of the redundancy issue in health and medical corpora.   

In our future work, we will develop dynamic and hierarchal topic models using fuzzy perspective. In addition, FLSA will be applied on social media data to track public opinions and will be used for online review and SMS spam detection \cite{karami2015online,karami2014improving}.

	\bibliography{refrence}

\vspace{2mm}

\textbf{Amir Karami} is an Assistant Professor in the School of Library and Information Science and a Faculty Associate at the Arnold School of Public Health at the University of South Carolina. He is currently working on developing fuzzy text mining techniques and their applications in medical, health, and social science. \\

\textbf{Aryya Gangopadhyay} is a Professor and the Chair of Information Systems Department at the University of Maryland Baltimore County. His research interests are in the areas of databases and data mining. Currently, he is focused on privacy preserving data mining, spatio-temporal data mining, and data mining for health informatics. \\

\textbf{Bin Zhou} is an Assistant Professor in the Department of Information Systems at the University of Maryland, Baltimore County. His recent research projects include designing efficient search techniques on large-scale data, detecting malicious activities that manipulate the Web search results, and protecting users privacy in large social networks. \\

\textbf{Hadi Kharrazi} is an Assistant Professor in the Johns Hopkins Bloomberg School of Public Health with a joint appointment at the Johns Hopkins School of Medicine. His research interest is in contextualizing CDSS in PHI platforms to be utilized at different HIT levels of managed care such as EHR platforms or consumer health informatics solutions. \\

	
		
		
		
\end{document}